\documentclass{article}
\usepackage{spconf,amsmath,graphicx}
\usepackage{url}
\usepackage{booktabs,threeparttable}
\usepackage{standalone}
\usepackage{multirow}
\usepackage{amssymb}

\usepackage[dvipsnames]{xcolor}

\newcommand{\RNum}[1]{\uppercase\expandafter{\romannumeral #1\relax}}

\title{
Sequence-to-sequence Automatic Speech Recognition \\
with word embedding regularization and fused decoding \\
}
\name{Alexander H. Liu$^{1}$\quad Tzu-Wei Sung$^{2}$\quad Shun-Po Chuang$^{1}$\quad Hung-yi Lee$^{1}$\quad Lin-shan Lee$^{1}$\thanks{Code available at \footnotesize{\protect\url{https://github.com/Alexander-H-Liu/End-to-end-ASR-Pytorch}}}}
\address{$^{1}$College of Electrical Engineering and Computer Science, National Taiwan University\\
         $^{2}$Department of Computer Science and Engineering, University of California, San Diego\\
         \small{\texttt{\{r07922013, b03902042, f04942141, hungyilee, lslee\}@ntu.edu.tw}
         }}
\begin{document}
%
\maketitle
\begin{abstract}
In this paper, we investigate the benefit that off-the-shelf word embedding can bring to the sequence-to-sequence (seq-to-seq) automatic speech recognition (ASR).
We first introduced the word embedding regularization by maximizing the cosine similarity between a transformed decoder feature and the target word embedding.
Based on the regularized decoder, we further proposed the fused decoding mechanism.
This allows the decoder to consider the semantic consistency during decoding by absorbing the information carried by the transformed decoder feature, which is learned to be close to the target word embedding.
Initial results on LibriSpeech demonstrated that pre-trained word embedding can significantly lower ASR recognition error with a negligible cost, and the choice of word embedding algorithms among Skip-gram, CBOW and BERT is important.
\end{abstract}

\section{Introduction}
\label{sec:intro}

End-to-end automatic speech recognition (ASR) systems~\cite{graves2013speech} have shown great success thanks to the fast advance of deep learning technologies.
With considerable quantities of parameters and training speech-transcription pairs, end-to-end ASR was proven to achieve performance comparable~\cite{amodei2016deep,prabhavalkar2017comparison} to (or even better~\cite{chiu2017state}  than) conventional ASR~\cite{jelinek1976continuous}.

However, the huge demand for fully annotated audio training data often makes end-to-end ASR non-applicable under different scenarios such as low-resourced languages or domain-specific speech recognition.
Furthermore, the superfluous capacity of deep ASR networks often leads to over-fitting problems, yielding a significant performance degradation when the condition of input speech differs.

Unlike annotated speech, text data can be collected much more easily.
To this end, substantial effort had been made to utilize pure text data for end-to-end ASR training.
Typical examples include the classic language model rescoring methods~\cite{mikolov2010recurrent,chorowski2016towards}, training an additional text-to-speech model to produce pseudo paired speech-text data~\cite{hayashi2018back,tjandra2017listening,baskar2019self}, and achieving adversarial learning by training an extra criticizing language model~\cite{liu2019adversarial}.
These methods pointed out the potential help that pure text data can offer in improving recognition performance.

Different from the previous works, in this paper we choose to utilize the contextual information distilled from text data via Word Embedding~\cite{mikolov2013linguistic}, which has been widely used in natural language processing nowadays.
With an adequate quantity of text data, word embedding ranging from Skip-gram, Continuous Bag Of Words (CBOW)~\cite{mikolov2013efficient,mikolov2013distributed} to Bidirectional Encoder Representation from Transformer (BERT)~\cite{devlin2018bert} can be learned in an unsupervised manner.
The learned embedding vectors can be considered as a continuous representation of the discrete words carrying semantic relationships in the latent vector space.

In this work, we seek to improve end-to-end ASR with word embedding learned from text-only data.
We choose to adopt word embedding because off-the-shelf word embedding carrying semantic information learned from a vast amount of text can be easily obtained.
We target at the family of sequence-to-sequence (seq-to-seq) ASR, where an autoregressive decoder was generally adopted to predict the transcription corresponding to the input speech.
We studied the benefit that word embedding can bring to this type of seq-to-seq ASR model and highlighted our contributions below:
\begin{itemize}
  \item We show that pre-trained word embedding can serve as an additional target for seq-to-seq ASR regularization. 
  \item We propose a fused decoding mechanism to utilize the word embedding during decoding.
  \item For both proposed methods, we evaluated the improvements made by different word embedding algorithms under different scenarios.
\end{itemize}

\begin{figure*}[ht]
\centerline{\includegraphics[width=14cm]{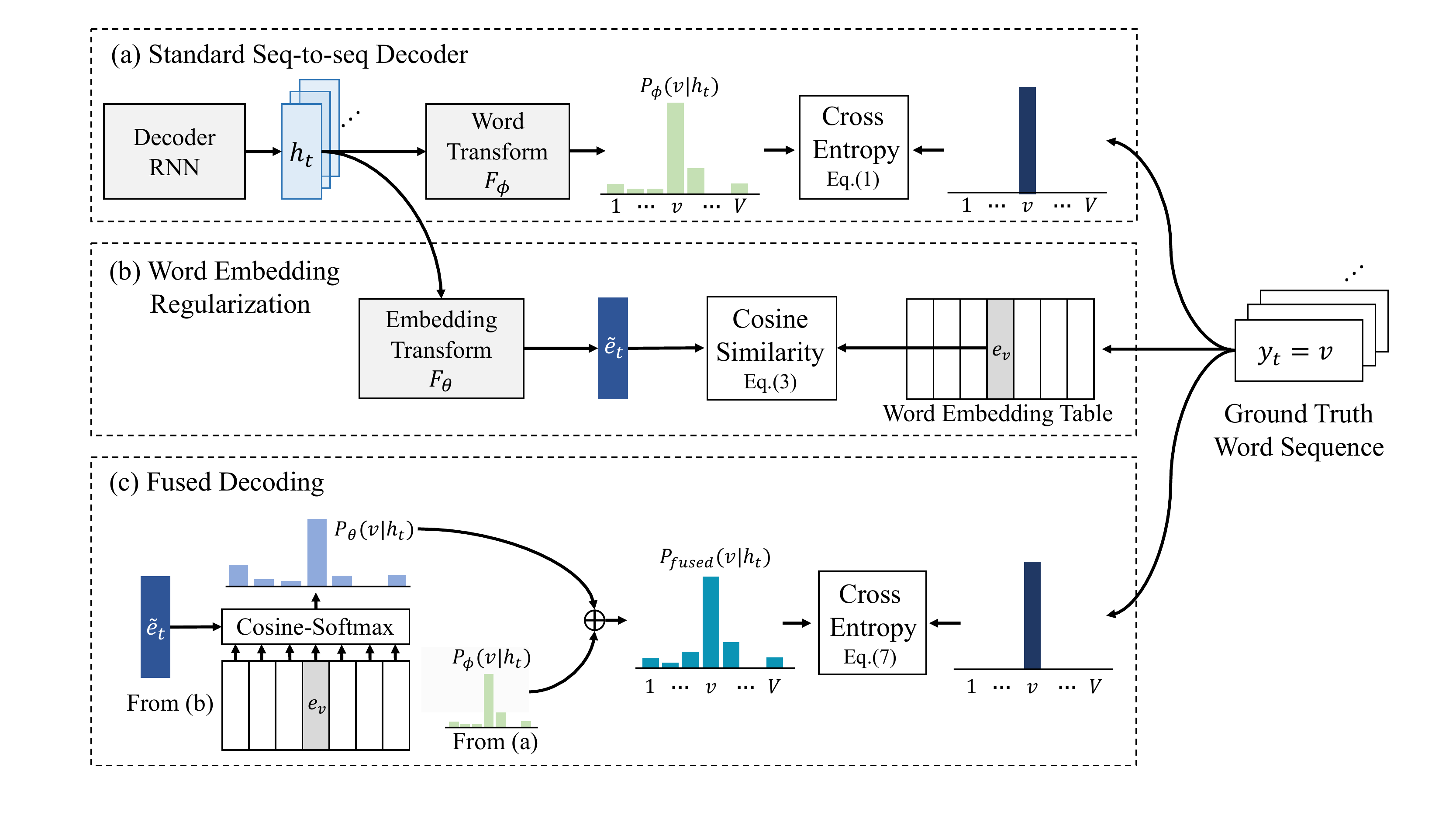}}
\vspace{-4pt}
\caption{Overview of the proposed approaches.
For each word-level time step $t$, the latent representation $h_t$ for the input speech obtained on the decoder and its corresponding target word $y_t$ is used.
(a) Standard seq-to-seq decoding: $h_t$ is transformed into a distribution $P_{\phi}(v|h_t)$ over the word vocabulary by a simple feed forward network $F_\phi$;
(b) Word embedding regularization: maximizing the cosine similarity between the projected feature $F_\theta(h_t)=\widetilde{e_{t}}$ and the target word embedding $e_{y_t}$;
(c) Fused decoding: a new probability distribution $P_\theta(v|h_t)$ based on $F_\theta (h_t)=\widetilde{e_{t}}$ (obtained by the Cosine-Softmax operation defined in Eq.~(\ref{eq:p_emb}) with the word embedding table) is fused with $P_\phi(v|h_t)$ in (a).}

\label{fig:overview}
\vspace{-7pt}
\end{figure*}


\section{Seq-to-seq ASR with word embedding}
\vspace{-3pt}
\label{sec:method}

This work is based on the existing paradigm of end-to-end sequence-to-sequence (seq-to-seq) ASR~\cite{chan2016listen}.
For each input utterance, the frame-level acoustic features are first extracted with a sequence encoder.
These extracted features are then decoded by a sequence decoder, producing a word-level latent representation $h_t$, where $t$ is the word-level time index.
As illustrated in Fig.~\ref{fig:overview}(a), each latent representation $h_t$ is transformed into a probability distribution $P_\phi(v|h_t)$ over the vocabulary of all considered words $v$ given $h_t$.
This is done by a word transformation $F_\phi$, which includes a simple feed-forward network plus a softmax activation.
The objective function of seq-to-seq ASR is then simply the log-likelihood function (or equivalently cross entropy)
\begin{equation}
    \label{eq:likelihood}
    L_{\text{asr}}(h_t, y_t) = - \log P_\phi(y_t|h_t)
\end{equation}where $y_t$ is the ground truth word at time step $t$ and $P_\phi(y_t|h_t)$ is $P_\phi(v|h_t)$ when $v=y_t$ given the decoder representation $h_t$.
Such encoder-decoder based ASR had been shown effective for transcribing speech given a sufficient amount of training data~\cite{chiu2017state}. The other parts of ASR including the encoder are left out in the above description for simplicity.

Approaches proposed in this paper included word embedding regularization as illustrated in Fig.~\ref{fig:overview}(b), and the fused decoding as shown in Fig.~\ref{fig:overview}(c).
Since these approaches only require a decoder that can generate output text sequence autoregressively, it can be equally applied to any encoder-decoder based ASR.
We now detail the proposed methods below.


\subsection{Word Embedding Regularization}
\label{subsec:reg}

Inspired by recent work on machine translation~\cite{unanue2019rewe}, we introduce the word embedding regression for regularizing seq-to-seq ASR.
Since word embedding carries distilled semantic information of words~\cite{mikolov2013linguistic}, the regularization term is to maximize the cosine similarity between a mapped latent feature $F_\theta(h_t)$ and the fixed word embedding $e_{y_t}$ of the corresponding ground truth $y_t$.

To be more specific, the latent representation $h_t$ produced in the seq-to-seq decoder as shown in Fig.~\ref{fig:overview}(a) is first projected to $R^D$ with a small network $F_\theta$, the embedding transform, 
\begin{equation}
\label{eq:proj}
\widetilde{e_t} = F_\theta(h_t),    
\end{equation}
as illustrated in Fig.~\ref{fig:overview}(b), where $D$ is the dimensionality of the word embedding considered.

Next, with the pre-trained word embedding vectors $E=\{e_1,e_2,...,e_V | e_i \in \mathbb{R}^D\}$  of vocabulary size $V$, the word embedding regularization term is defined as
\begin{equation}
\label{eq:reg}
    L_\text{reg}(\widetilde{e_t}, e_{y_t}) = 1 - \cos(\widetilde{e_t},e_{y_t}) =  {1 - \frac{\widetilde{e_t} \cdot e_{y_t}}{{\|\widetilde{e_t}\|}_2 \cdot {\|e_{y_t}\|}_2}},
\end{equation}
where $e_{y_t} \in E$ with $y_t$ being the ground truth word at time $t$. 
So the goal is to have $F_\theta(h_t)$ close to $e_{y_t}$. 

Combining Eq.~(\ref{eq:likelihood})(\ref{eq:proj})(\ref{eq:reg}), the complete objective function of ASR with word embedding regularization is defined as the following:
\begin{equation}
\label{eq:full_reg}
    L = \sum_{t}(L_{\text{asr}}(h_t,y_t) + \lambda  L_{\text{reg}}(F_\theta(h_t), e_{y_t})),
\end{equation}
where $\lambda$ controls the intensity of regularization.
Note that the second term serves as a regularization for training only, since $F_\theta(h_t)$ is never referenced in decoding.
It is also worth mentioning that the computational cost of such regularization is negligible given $F_\theta$'s simplicity compared against the seq-to-seq ASR and the fixed embedding table $E$  without being updated throughout the training.


\subsection{Fused Decoding}
\label{subsec:joint}

With the embedding regularization introduced in Sec.~\ref{subsec:reg}, $\widetilde{e_t}$ is expected to be similar to its corresponding target word embedding, but not used in decoding.
Here we further pursue to utilize such characteristic of $\widetilde{e_t}$ and propose the fused decoding mechanism, as illustrated in Fig.~\ref{fig:overview}(c).

Based on the cosine similarity in Eq.~(\ref{eq:reg}), we first defined the Cosine-Softmax operation to obtain a probability distribution $P_\theta(v|h_t)$ over all words $v$ in vocabulary with size $V$ given the regularized decoder latent representation $\widetilde{e_t}$ and the embedding table $E = \{e_1,e_2,...,e_V\}$ to be
\begin{equation}
    \label{eq:p_emb}
    P_\theta(v|h_t) = \frac{\exp(\cos (\widetilde{e_t} , e_{v}) / \tau)}{\sum_{k \in V} \exp(\cos ({\widetilde{e_t} , e_k})/ \tau)},
\end{equation}
where $\widetilde{e_t}$ is $F_\theta(h_t)$ from Eq.~(\ref{eq:proj}), $e_k$ is the word embedding of word $k$ and $\tau$ is the temperature of softmax function.
Since the magnitude of embedding was not accounted by cosine similarity, the choice of $\tau$ will dominate the sharpness of $P_\theta$.

Combining Eq.~(\ref{eq:p_emb}) with the original probability $P_\phi(v|h_t)$ in Fig.~\ref{fig:overview}(a), the fused probability over the vocabulary set $V$ at time $t$ is defined as
\begin{equation}
    \label{eq:p_fuse}
    P_\text{fused}(v|h_t) = (1-\lambda_f) P_\phi(v|h_t) + \lambda_f P_\theta(v|h_t)
,\end{equation}
where the first term is from Fig.~\ref{fig:overview}(a), the second term is from Eq.~(\ref{eq:p_emb}) and $\lambda_f$ is the weight of fusion.

To construct the complete objective function of ASR with fused decoding, we first replace the original probability distribution $P_\phi(v|h_t)$ in Eq.~(\ref{eq:likelihood}) by the fused probability $P_\text{fused}(v|h_t)$ in Eq.~(\ref{eq:p_fuse}) to have
\begin{equation}
    \label{eq:l_fuse}
    L_\text{fused}(h_t,y_t) = -\log P_\text{fused}(y_t|h_t),
\end{equation}
which is very similar to Eq.~(\ref{eq:likelihood}), and then rewrite Eq.~(\ref{eq:full_reg}) to obtain the complete objective function including the regularization term:
\begin{equation}
\label{eq:full_fuse}
    L = \sum_{t}(L_\text{fused}(h_t,y_t) + \lambda  L_\text{reg}(F_\theta(h_t), e_{y_t})).
\end{equation}

Unlike the regularization method introduced in Sec.~\ref{subsec:reg}, this fused decoding allows the mapped features $\widetilde{e_t}$ to modify the output distribution of ASR based on the similarity to the target word embeddings.
We restate that similar to Sec.~\ref{subsec:reg}, the word embedding table $E$ was fixed without updating throughout the training and no additional learnable parameter was introduced for the proposed fused decoding.

\begin{table*}[ht]
\small
\centering
\begin{threeparttable}

\caption{Speech recognition word error rate (\%) on LibriSpeech with 460 hours training data (high resource setting).}

\begin{tabular}{c| c c | c | c c c c | c c c c }
\toprule
\multirow{3}{*}{Index} &Word& Fused & Word  & \multicolumn{4}{c}{(\RNum{1}) WER (\%)}  & \multicolumn{4}{|c}{(\RNum{2}) WER (\%), w/ RNN-LM} \\
& Embedding  &   & Embedding      & \multicolumn{2}{c}{Clean} & \multicolumn{2}{c}{Other} & \multicolumn{2}{|c}{Clean} & \multicolumn{2}{c}{Other} \\
& Regularization  & Decoding  & Type    & Dev & Test & Dev & Test & Dev & Test & Dev & Test \\ \hline
(a) Baseline & - & - & - & 13.41 & 14.28 & 37.95 & 40.03 & 11.18 & 12.37 & 36.84 & 38.96 \\ \hline
(b) & \checkmark  & - & \multirow{2}{*}{Skip-gram} & 12.23 &  \textbf{12.78} & 36.72 & 38.54 & 10.79 & \textbf{11.62} & 35.72 & 38.13\\
(c) & \checkmark  & \checkmark &                   &  12.36 & 13.06& 37.02 & 38.69  & 10.62 & 11.95& 35.99 & \textbf{38.00} \\ \hline
(d) & \checkmark  & - & CBOW &  12.67 & 13.08 & 38.11 & 39.26 & 11.50 & 12.19 & 37.92 & 39.59\\
(e) & \checkmark & - & BERT & \textbf{11.84} & 12.82 & \textbf{36.12} & \textbf{38.38} & \textbf{10.56} & 11.79 & \textbf{35.54} & 38.01 \\ 
\bottomrule
\end{tabular}
\vspace{-5pt}
\label{table:high}
\end{threeparttable}
\end{table*}

\section{Experiments}
\label{sec:exp}
\vspace{-4pt}
\subsection{Experimental Setup}
\label{subsec:setting}

Experiments were performed on LibriSpeech~\cite{panayotov2015librispeech} with two different settings:
1) The high resource setting with 460 hours of fully annotated clean speech and pure text data from the transcriptions of 960 hours.
2) The low resource setting with 100 hours of annotated clean speech and the transcriptions of 460 hours clean speech to train the word embedding.

For the seq-to-seq ASR, the encoder was fixed as a 4-layer convolution network~\cite{hori2017advances} and 5-layer bidirectional LSTM with a hidden size of 512 on each direction.
The decoder was a single layer LSTM with a size of 512 and used location-aware attention~\cite{chorowski2015attention}.
The word transform layer $F_\phi$ was a single layer linear transform followed by a softmax activation.
The embedding transform layer $F_\theta$ was a two-layer fully connected neural network.
Text data was encoded into a sequence of subword units out of a vocabulary set with size  5000.
The embedding for each subword token with a fixed dimensionality of 256 was obtained through a fast off-the-shelf word embedding extraction toolkit FastText~\cite{bojanowski2016enriching}.
Besides Skip-gram and CBOW provided by FastText, we also involved a pre-trained 12 layer BERT~\cite{wolf2019transformers} which was fine-tuned on LibriSpeech.
For BERT, the averaged embedding over all layers for each token was used as the regularization target.
For high resource experiments, we fixed $\lambda$ in Eqs.~(\ref{eq:full_reg})(\ref{eq:full_fuse}) to 10 and both $\tau$~/~$\lambda_f$ in Eq.~(\ref{eq:p_fuse}) to 0.1.
For the low resource experiments, $\tau$ was set to 0.02 and multitask learning of CTC~\cite{graves2006connectionist,kim2017joint} was performed on encoder during training while the CTC output was ignored during test.
Beam search decoding with the beam size set of 20 is used for inference.
All models were selected based on their word error rate on the clean development set.
We reported the averaged word error rate of each method over different random initialization for objective evaluation.
For more detailed setup, please refer to our implementation.

\vspace{-10pt}

\subsection{Results on High Resource ASR}
\label{subsec:high_exp}

With 460 hours of training data, all results are shown in Table~\ref{table:high}.
We also combined our method with joint RNN-LM decoding~\cite{hori2017advances} where the RNN-LM was separately trained on the identical text corpus used by word embedding (columns (\RNum{2}) v.s. (\RNum{1})).

\noindent \textbf{Recognition Error Rate with Skip-gram} We first compare the word embedding regularization (row(b)) or plus fused decoding (row(c)) against the baseline (Fig.~\ref{fig:overview}(a), row( a)) with Skip-gram for embedding.
We observe a significant improvement made by both proposed methods (rows(b)(c) v.s. (a)) and a consistent improvement when the testing environment shift from clean to noisy (column "Clean" v.s. "Other")
The extra gain brought by fused decoding on top of regularization (row(c) v.s. (b)) was not clear here and we'll see its value later in the low resource setting.

We also discovered that our methods were compatible to joint language model decoding (columns(\RNum{2}) v.s. (\RNum{1})), but with less improvement made by RNN-LM to our proposed methods (row(b)(c)) considering the significant improvement RNN-LM brought to the baseline (row(a)).
This indicates that the proposed methods poured some semantic-level information to ASR, part of which was similar to language modeling, while others beyond.

\noindent \textbf{Word Embedding Selection}
The result for word embedding extracted by CBOW~\cite{mikolov2013efficient} and BERT~\cite{devlin2018bert} are in rows(d)(e).
Fused decoding cannot be used with BERT due to the bi-directional nature of BERT.
We see CBOW was not as good as Skip-gram for regularization (row(d) v.s.(b)), while the contextual embedding from BERT had generally improved the ASR most significantly (row(e) v.s. (b)(c)(d)), although this no longer holds when an additional RNN-LM is available (columns(\RNum{2}) v.s. (\RNum{1})).
This is consistent to the above statement that the impact of word embedding on ASR somehow overlapped with language modeling, since BERT was also known as a masked language model.
Nevertheless these results demonstrated the effectiveness of our proposed methods when a strong embedding is available.

\vspace{-7pt}
\subsection{Results on Low Resource ASR}
\label{subsec:low_exp}

With only 100 hrs of annotated data, the results are in Table~\ref{table:low}.
"w/ LM" for column(\RNum{2}) indicates joint RNN-LM decoding.

\noindent \textbf{Performance Study} We first compare our proposed methods against the baseline, and found consistent improvements when only limited training data was available (rows(b)(c) v.s. (a), columns(\RNum{1})(\RNum{2})).
We highlight that the fused decoding consistently made extra improvement on top of that by regularization (rows(b) v.s. (a)), although all improvement made here were not as high as that obtained with the high resource setting in Table~\ref{table:high}.

\noindent \textbf{Comparing Against Semi-supervised Methods} We also listed the performance obtained with the same setting reported by prior works (referred to as "semi-supervised") for comparison.
Our word embedding regularization surpassed the back-translation data augmentation method~\cite{hayashi2018back} (row(d)) yet still performed worse than the adversarial training method~\cite{liu2019adversarial} (row(e)).
With fused decoding, we further narrowed the gap.
However, it is worth mentioning that all the semi-supervised methods listed in Table~\ref{table:low} required ASR counterpart training (a text-to-speech model~\cite{baskar2019self,hayashi2018back} or a discriminator ~\cite{liu2019adversarial}) to optimize the performance at the price of higher computational resource.
But our methods add nearly no cost \footnote{As a reference, the time  spent on pre-training word embedding took approximately 2\% out of the total training time.} in training.

\begin{table}[t]
\small
\centering
\begin{threeparttable}

\caption{\small Speech recognition word error rate (\%) on LibriSpeech with 100 hours training data (low resource setting). "w/ LM" indicated jointly RNN-LM decoding~\cite{hori2017advances}.}

\begin{tabular}{l| c c | c c }
\toprule
 \multirow[c]{4}{*}{Methods} & \multicolumn{2}{c|}{(\RNum{1})} &\multicolumn{2}{c}{(\RNum{2})}\\
&\multicolumn{2}{c|}{ \multirow[c]{2  }{*}{WER(\%)}} &  \multicolumn{2}{c}{ WER (\%)}  \\ 
& & &  \multicolumn{2}{c}{ w/ LM}  \\ 
                                        &   Dev    &   Test    &   Dev   &   Test \\ \hline
(a) Baseline                                &  21.7     &   22.3    &   19.1  & 19.9     \\
(b) Proposed Regularization                     &   20.4    &   21.5    &   18.5  & 19.3 \\ 
(c) Proposed Fused Decoding                 &   20.2       &   21.2       &   18.2     & 19.1     \\ \hline\hline
Semi-supervised Methods                  &           &           &         &         \\ \hline
(d) Back-translation~\cite{hayashi2018back} &   23.5    &   23.6    &   21.6  & 22.0    \\
(e) Criticizing-LM~\cite{liu2019adversarial} &   19.1    &   19.2    &   17.1  & 17.3    \\
(f) Semi-ASR w/ TTS~\cite{baskar2019self}   &   -       &   17.9    &   -     & 17.0    \\

\bottomrule
\end{tabular}

\label{table:low}
\end{threeparttable}
\end{table}

\section{Conclusion}
\label{sec:conclusion}

In this paper, we proposed two methods to utilize word embedding on seq-to-seq ASR.
Word embedding regularization can enforce the latent decoder features to be closely related to the corresponding target word embedding, which can be further used to enhance the ASR output with fused decoding.
These methods were tested and found useful under different scenarios and different choice of word embedding.
Most importantly, the proposed methods can serve as plug-in enhancers for any seq-to-seq ASR with a small computational price.

\newpage
\vfill\pagebreak
\bibliographystyle{IEEEbib}
{\small\bibliography{strings,refs}}
\end{document}